\documentclass[letterpaper, 10 pt, conference]{ieeeconf}

\usepackage{graphicx}
\usepackage{amsmath}
\usepackage{amssymb}
\usepackage{csquotes}
\usepackage{url}
\usepackage{hyperref}
\usepackage{multirow}
\usepackage[table]{xcolor}
\usepackage{algorithm}% http://ctan.org/pkg/algorithms
\usepackage{tabularray}
\usepackage{algpseudocode} %http://ctan.org/pkg/algorithmicx\usepackage{array,multirow,graphicx}
\usepackage[nolist]{acronym} % acronyms by the \ac{label} command
\let\labelindent\relax
\usepackage[inline]{enumitem} % inline enumerate
\usepackage{cite}
\graphicspath{ {imgs/} }
\usepackage{siunitx}
\usepackage{censor}
\usepackage{booktabs}

\newacro{dnn}[DNN]{Deep Neural Network}
\newacro{fcn}[FCN]{Fully Convolutional Network}
\newacro{sdf}[SDF]{Signed Distance Function}
\newacro{cnn}[CNN]{Convolutional Neural Network}
\newacro{gnn}[GNN]{Graph Neural Network}
\newacro{dl}[DL]{Deep Learning}
\newacro{ml}[ML]{Machine Learning}
\newacro{gpis}[GPIS]{Gaussian Process Implicit Surface}
\newacro{mlp}[MLP]{Multi-Layer Perceptron}
\newacro{dof}[DoF]{Degree of Freedom}

\newacro{ai}[AI]{Artificial Intelligence}
\newacro{llm}[LLM]{Large Language Model}
\newacro{hri}[HRI]{Human-Robot Interaction}
\newacro{phri}[pHRI]{Physical Human-Robot Interaction}
\newacro{shri}[sHRI]{Social Human-Robot Interaction}

\newacro{slam}[SLAM]{Simultaneous Localization and Mapping}
\newacro{pf}[PF]{Particle Filter}
\newacro{kf}[KF]{Kalman Filter}

\newacro{ros}[ROS]{Robot Operating System}
\newacro{gui}[GUI]{Graphical User Interface}
\newacro{urdf}[URDF]{Unified Robot Description Format}

\newacro{ik}[IK]{Inverse Kinematics}

\newacro{rrmc}[RRMC]{Resolved-Rate Motion Control}
\newacro{yarp}[YARP]{Yet Another Robot Platform}

\newacro{eit}[EIT]{Electrical Impedance Tomography}
\newacro{gp}[GP]{Gaussian Process}

\newacro{ft}[F/T]{Force/Torque}
\newacro{pl}[PL]{Pain Level}
\newacro{nwr}[NWR]{Nociceptive Withdrawal Reflex}

\newacro{ur}[UR]{Uniform Reflex}
\newacro{lr}[LR]{Location-dependent Reflex}
\newacro{cr}[CR]{Cartesian Reflex}

\newcommand{\equationref}[1]{\hyperref[#1]{Eq.~\ref*{#1}}}
\newcommand{\figref}[1]{\hyperref[#1]{Fig.~\ref*{#1}}}
\newcommand{\tabref}[1]{\hyperref[#1]{Table~\ref*{#1}}}
\newcommand{\secref}[1]{\hyperref[#1]{Section~\ref*{#1}}}
\newcommand{\algoref}[1]{\hyperref[#1]{Alg.~\ref*{#1}}}

\newcommand{\etal}[1]{#1 \textit{et al.}}

\newif\ifuseVspace

\useVspacefalse % true to use vspaces

\usepackage{etoolbox}

\newif\ifcensoractive
\censoractivetrue 
\apptocmd{\StopCensoring}{\censoractivefalse}{}{}
\apptocmd{\RestartCensoring}{\censoractivetrue}{}{}

\newcommand{\censorurl}[1]{%
  \ifcensoractive
    \blackout{#1}
  \else
    \url{#1}
  \fi
}

\StopCensoring

 \usepackage[firstpageonly=true]{draftwatermark}

\SetWatermarkAngle{0}
\SetWatermarkColor{black}
\SetWatermarkLightness{0.5}
\SetWatermarkFontSize{9pt}
\SetWatermarkVerCenter{30pt}
\SetWatermarkText{\parbox{30cm}{%
\centering This work has been submitted to the IEEE for possible publication. \\
\centering Copyright may be transferred without notice, after which this version may no longer be accessible.
}}

\title{Design and Human Evaluation of Tactile Withdrawal Reflexes for a Skin-Covered Robot Arm}
\xblackout{
\author{Laura Babayeva, Lukas Rustler, and Matej Hoffmann %
\thanks{Laura Babayeva, Lukas Rustler and Matej Hoffmann are with the Department of Cybernetics, Faculty of Electrical Engineering, Czech Technical University in Prague,
{\tt\small lukas.rustler@fel.cvut.cz, matej.hoffmann@fel.cvut.cz.}}
 \thanks{This work was co-funded by the European Union under the project Robotics and Advanced Industrial Production (reg. no. CZ.02.01.01/00/22\_008/0004590). L.R. was additionally supported by the Czech Science Foundation (GA ČR) under project No. 26-22606S and Grant Agency of the Czech Technical University in Prague, grant No. SGS26/075/OHK3/1T/13. M.H. was additionally supported by the Czech Science Foundation (GA ČR), project no. 25-18113S.}}}

\IEEEoverridecommandlockouts
\begin{document}

\maketitle

%%%%%%%%%%%%%%%%%%%%%%%%%%%%%%%%%%%%%%%%%%%%%%%%%%%%%%%%%%%%%%%%%%%%%%%%%%%%%%%%
% 2250 Characters for Humanoids -> current version is ~1550
\begin{abstract}
Nociception is a protective biological mechanism that links harmful stimulation to a reaction, such as rapid withdrawal. This paper investigates artificial nociception for a collaborative robotic arm with whole-body tactile sensing. We present a complete pipeline that maps pressure changes from sensitive skin on a robot manipulator to bio-inspired withdrawal motions. The system first converts skin pressure into a scalar pain gain using a nonlinear continuous model. We compare three reflexes: (i) uniform reflex moves four robot joints by a fixed amount, whereby the withdrawal is approximated by a stereotypical movement of the arm "toward the base",  independent of where the robot was touched; (ii) biologically motivated location-dependent joint-space withdrawal derived from human withdrawal reflex characteristics; (iii) Cartesian space withdrawal along the surface normal of the contacted skin pad. All behaviors are integrated in a finite-state reflex controller that interrupts the task, executes the withdrawal, and returns to a pre-contact pose. A user study with 15 participants compared the three strategies using Godspeed questionnaire subscales, custom perceived-naturalness and safety items, forced-choice comparisons, and qualitative feedback. Interestingly, participants rated more highly the stereotypical uniform reflex behavior over one or both competitors on the anthropomorphism, animacy, and likeability Godspeed subscales and on the Naturalness and Realism custom scale. When asked to compare the conditions, the uniform reflex was scored best in "felt safest", "most human-like", and "most natural". This suggests that predictability of the robot behavior is key for user acceptance. The Cartesian reflex was judged the most appropriate reaction to touch. The bio-inspired reflex did not lead any evaluated measure. This may be partly attributed to the embodiment gap between the robot arm and human arm and participants having different expectations from a robot manipulator.
\end{abstract}

\section{Introduction}

More than a century ago, Karel Čapek in his play R.U.R. (Rossum's Universal Robots) imagined robots that, because they felt almost no bodily pain, failed to protect themselves from damage. Dr. Gall's remedy was to add pain sensitivity as an automatic protective mechanism. The addition of pain and irritability could be interpreted as a step by which Rossum's robots began to care about their bodies and, ultimately, themselves. Setting aside the play's dystopian trajectory, the underlying engineering insight remains strikingly current: robots that share physical space with people require distributed tactile sensitivity and rapid, intelligible protective responses. In contemporary terms, the relevant construct is artificial nociception—--the detection and encoding of potentially damaging stimulation—--rather than subjective pain. This paper investigates how such signals should be translated into withdrawal behavior on a skin-covered robot arm.

Humanoid and human-centered robots increasingly operate in cluttered environments and in close physical contact with people. In such settings, safety cannot rely only on distance separation or emergency stops on contact. The robot must also interpret collisions and react in a way that is safe for both humans and robots. Biological pain provides a useful design metaphor for this problem. In humans, nociception is a protective sensorimotor mechanism: potentially damaging stimulation can trigger rapid withdrawal before the stimulus causes tissue damage.

\begin{figure}[t]
    \centering
     \includegraphics[width=0.99\linewidth]{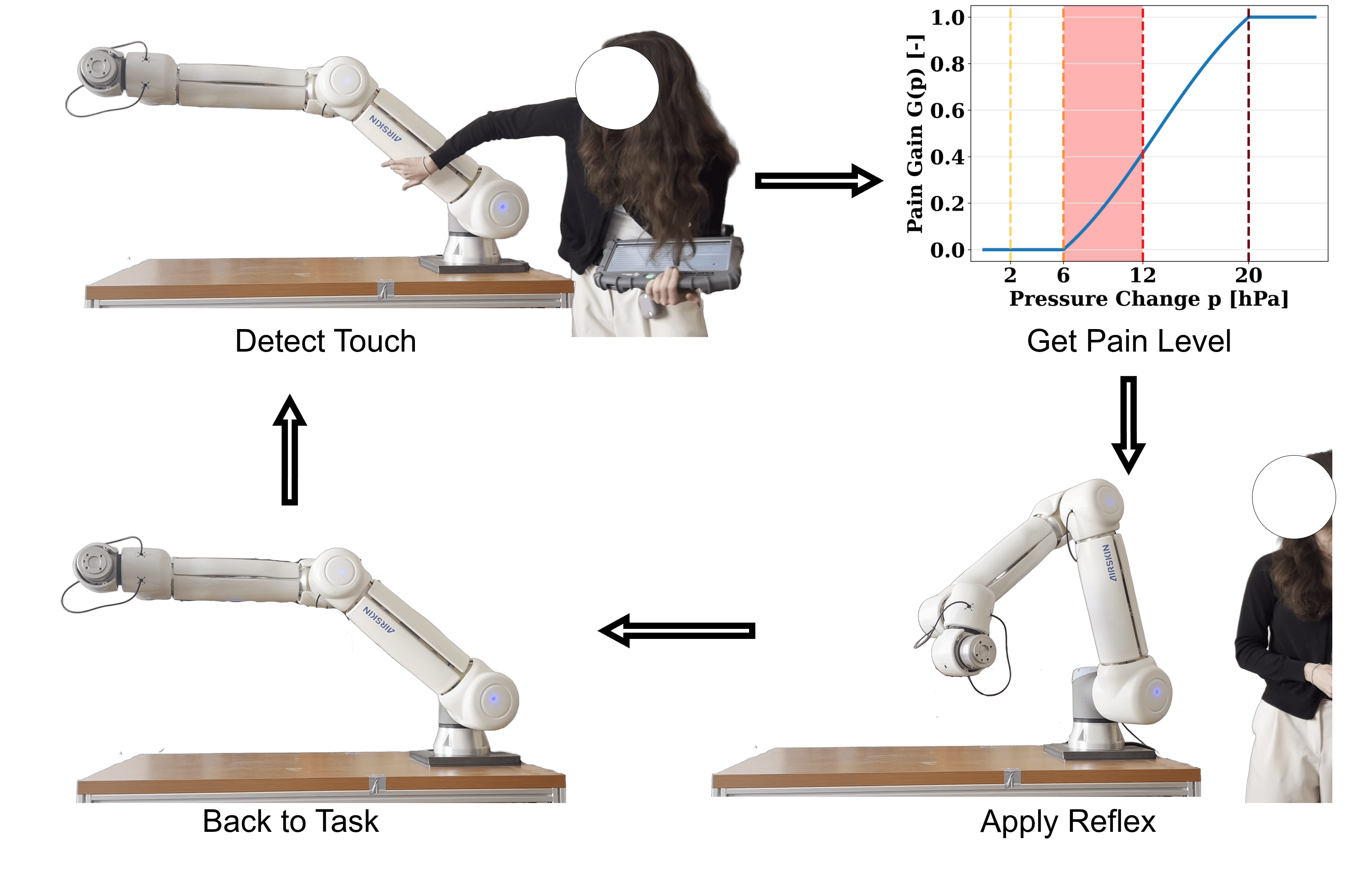}
    \vspace{-2em}
    \caption{Schematic operation of the pipeline. The robot performs a task. If touch is detected, a pain gain $G(p)$ is estimated using the change in pressure $p$. A reflex is executed based on the calculated pain gain---if the touch force is gradually increasing or there is a new contact during the reflex, the reflex can be adjusted. After the reflex is done, the robot returns to the position before contact and continues the task.}
    \label{fig:schema}
\end{figure}

Prior work already addresses several links in the tactile-to-reflex chain. Kuehn and Haddadin introduced an artificial robot nervous system that maps multimodal tactile input to nociceptive states and impedance-based reflex motion~\cite{kuehn2016artificial}. Vorndamme et al. developed a context-sensitive reflex engine that selects among task-preserving, task-relaxing, and task-abandoning post-contact maneuvers~\cite{Vorndamme2022RobotContactReflexes}. In parallel, Liu et al. reviewed neuromorphic hardware and processing architectures for large-area electronic skin~\cite{Liu2022NeuroInspired}, while Feng and Zeng proposed a spiking-network model of actual and anticipated robot injury and evaluated it in two real-robot tasks~\cite{Feng2022BrainInspired}. Thus, the missing element is not a complete tactile-to-reflex pipeline per se. The contribution of this work is the within-platform comparison of three withdrawal mappings using distributed tactile input, together with a human-centered evaluation of the resulting predictability, naturalness, and appropriateness.

This paper is based on a tactile UR10e manipulator covered with AIRSKIN\footnote{\url{https://www.airskin.io/}} pressure pads. We explore artificial nociception and ask: how should a robot arm withdraw when touched with different pressure levels at different body regions? This is important for \ac{hri} because the same movement can serve a functional safety role and a communicative role. A reflex that is mechanically protective but illegible may still feel unsafe; conversely, a reflex that is predictable and understandable may be preferred even when it is less faithful to biological joint recruitment. A schematic overview of our pipeline is available in Fig.~\ref{fig:schema}.

\begin{table*}[t]
\centering
\footnotesize
\caption{Overview of the NWR pathway in the upper limb, showing the relationship between arm regions, spinal segments, involved nerves and muscles, and resulting movements.}
\label{tab:nwr_upper_limb}

\begin{tblr}{
width=\textwidth,
colspec={
Q[l,1.5cm]
Q[c,1.5cm]
Q[l,2.1cm]
X[2.5,l]
X[2.8,l]
},
row{1} = {font=\bfseries},
rows = {valign=m},
rowsep = 2pt,
colsep = 4pt,
hline{1,Z} = {1.1pt, black},
hline{2} = {0.8pt, black},
hline{3,5,6,8,10} = {1.1pt, black},
hline{4,7,9} = {2,3}{0.3pt, black},
}

Body region &
Spinal segments~\cite{Polcaro2023BrachialPlexus} &
Nerves~\cite{kaiser2023anatomy, bayot2023anatomy} &
Dominant muscles~\cite{serrao2006kinematic} &
Resulting movement \\

Shoulder &
C5--C6 &
Auxiliary &
Activation of anterior deltoid; inhibition of posterior deltoid &
Shoulder anteflexion, shoulder abduction/adduction, internal rotation, and elbow flexion \\

Upper arm &
C5--C7 &
Musculocutaneous &
\SetCell[r=2]{m} Activation of biceps and brachioradialis; inhibition of triceps &
\SetCell[r=2]{m} Elbow flexion and minor shoulder flexion \\

&
C5--T1 &
Radial &
&
\\

Elbow &
C5--C7 &
Musculocutaneous &
Biceps brachii and brachioradialis activation &
Elbow flexion \\

Forearm &
C5--T1 &
Median and radial &
\SetCell[r=2]{m} Pronators or supinators depending on stimulus location &
\SetCell[r=2]{m} Forearm pronation/supination with elbow flexion \\

&
C8--T1 &
Ulnar &
&
\\

Wrist and hand &
C5--T1 &
Median and radial &
\SetCell[r=2]{m} Wrist flexor activation; extensor inhibition &
\SetCell[r=2]{m} Wrist adduction/abduction depending on stimulus location, with wrist and elbow flexion \\

&
C8--T1 &
Ulnar &
&
\\
\end{tblr}
\end{table*}

\textbf{Contributions.} We make the following contributions. First, we provide a summary of human \acf{nwr} pathways and pain levels, map it to the robotic manipulator, and define a biologically inspired nonlinear continuous model that converts pressure readings from the robotic skin to a pain gain factor. Second, we implement and compare three withdrawal controllers with a different level of biological inspiration and complexity of movement. Third, we evaluate how humans perceive these behaviors in direct interaction with the robot. The results of the study suggest that the most biologically inspired methods may not be the best reflected by humans. We provide data (the study questionnaire and responses), code, and video at \censorurl{https://humanoids-ctu.github.io/Nociceptive-Reflexes}.

%%%%%%%%%%%%%%%%%%%%%%%%%%%%%%%%%%%%%%%%%%%%%%%%%%%%%%%%%%%%%%%%%%%%%%%%%%%%%%%%

\section{Related Work}
\label{sec:related_work}

The human \ac{nwr} is a rapid polysynaptic spinal mechanism that protects the limbs from noxious stimuli without requiring cortical involvement~\cite{Derderian2019physiology, Bear2015Neuroscience}. The reflex coordinates motor responses by exciting flexor muscles and inhibiting extensors to mechanically distance the limb from harm~\cite{serrao2006kinematic}. Crucially, \ac{nwr} is not a stereotyped motion; it is highly context-dependent, modulated by stimulus location, intensity, and ongoing posture~\cite{HENRICH2022134, MassAlarie2023NociceptiveWR}. For example, proximal joints such as the elbow and shoulder are consistently recruited even when stimuli are applied distally to the hand~\cite{Peterson2014Withdrawal, serrao2006kinematic}. Table~\ref{tab:nwr_upper_limb} summarizes this biological organization in the upper limb, detailing the mapping between stimulus regions, spinal segments, and the resulting reflexive movements.

Biological pain encoding is continuous and nonlinear. Perceptual pain magnitude scales with stimulus intensity as a power function \cite{pain_intensity_processing, visual_analogue_scales}. Pain sensitivity also varies by body location due to differing densities of the nociceptors~\cite{alburquerque2018, spatial_acuity}. While continuous scaling is essential for biologically faithful modeling, discrete classifications of pain severity offer a highly interpretable baseline to validate different threshold behaviors. To establish this, we draw on the experimental conditions used by \etal{Serrao}~\cite{serrao2006kinematic}, who applied stimuli at 1 to 4 multiples of participant-specific pain thresholds. We use the responses observed as reference points for constructing an artificial nociceptive scale of pain level (PL) as follows: 
\begin{itemize}
    \item \textbf{PL 1 \& 2} No measurable movement. Interpreted as a non-threatening stimulus below the reflex activation threshold.
    \item \textbf{PL 3:} Mild reflex (Elbow flex: $\sim$2.3$^\circ$, Shoulder anteflex: $\sim$1.9$^\circ$). Moderate intensity where the reflex begins to activate.
    \item \textbf{PL 4:} Clear withdrawal (Wrist add: $\sim$5.2$^\circ$, Elbow flex: $\sim$6.5$^\circ$, Shoulder anteflex: $\sim$5.4$^\circ$). Strong stimulus that fully activates coordinated withdrawal.
\end{itemize}
The resulting robotic levels are an engineering abstraction and are not intended to be a direct equivalence between AIRSKIN pressure and human pain threshold.

In robotics, contacts can be perceived using intrinsic sensing relying for example on joint torque sensors (e.g., \cite{iskandar2024intrinsic}) or using large-area electronic skins (e.g., \cite{cheng2019comprehensive}). Distributed pressure sensors are highly capable when it comes to collision detection, isolation, and identification, which are necessary to choose and execute the appropriate reaction. Examples of using whole-body skins for p\ac{hri} span high-resolution skins to convert the touch signal into contact centers and estimating the corresponding wrench~\cite{Sun2025Beyond}, hierarchical framework with three levels of granularity for robot safety while navigating the environment~\cite{Jiang2024Hierarchical}, or calculating external forces to create robot reflex actions~\cite{Duong2025TactileReflex}. Rozlivek et al.~\cite{Rozlivek2025Harmonious} integrated tactile, proximity, and visual perception of humans as constraints for a whole-body humanoid robot controller. AIRSKIN---the sensor used in this work---was studied in \cite{svarny2022effect} to assess its contribution to reducing the impact forces after impact. In \cite{rustler2026adaptive}, it was used for collision detection and isolation, whereas the robot effective mass calculated online was used to estimate the impact forces and choose the appropriate reaction.   

Recent neuromorphic approaches have introduced artificial Robot Nervous Systems (aRNS) that emulate human nociceptors. Kuehn and Haddadin~\cite{kuehn2016artificial} developed a mechanical skin model where artificial neurons encode collision data into frequency-modulated spikes, triggering discrete heuristic reflex strategies based on the severity of the pain. Other works extend this bio-inspiration to the hardware level, using computational electronic skin (e-skin) with synaptic transistors for localized edge-processing of tactile stimuli~\cite{Liu2022Printed, yang_eskin}, or deploying Spiking Neural Networks (SNNs) to allow robots to learn pain associations and predict injury~\cite{Feng2022BrainInspired, ZENG2023100789}.

To safely integrate reflexive behaviors, specialized control architectures are required. \etal{Vorndamme}~\cite{Vorndamme2022RobotContactReflexes, Vorndamme2024SafeRobotReflexes} proposed a real-time reflex engine that overrides standard motion generators during collisions. By categorizing reactions into task-preserving, task-relaxing, and task-abandoning reflexes, the system provides a structured pipeline for managing contact severity.

%%%%%%%%%%%%%%%%%%%%%%%%%%%%%%%%%%%%%%%%%%%%%%%%%%%%%%%%%%%%%%%%%%%%%%%%%%%%%%%%
\section{Materials and Methods}

\subsection{Hardware and Software}
\begin{figure}[htb]
    \centering
    \includegraphics[width=0.575\columnwidth]{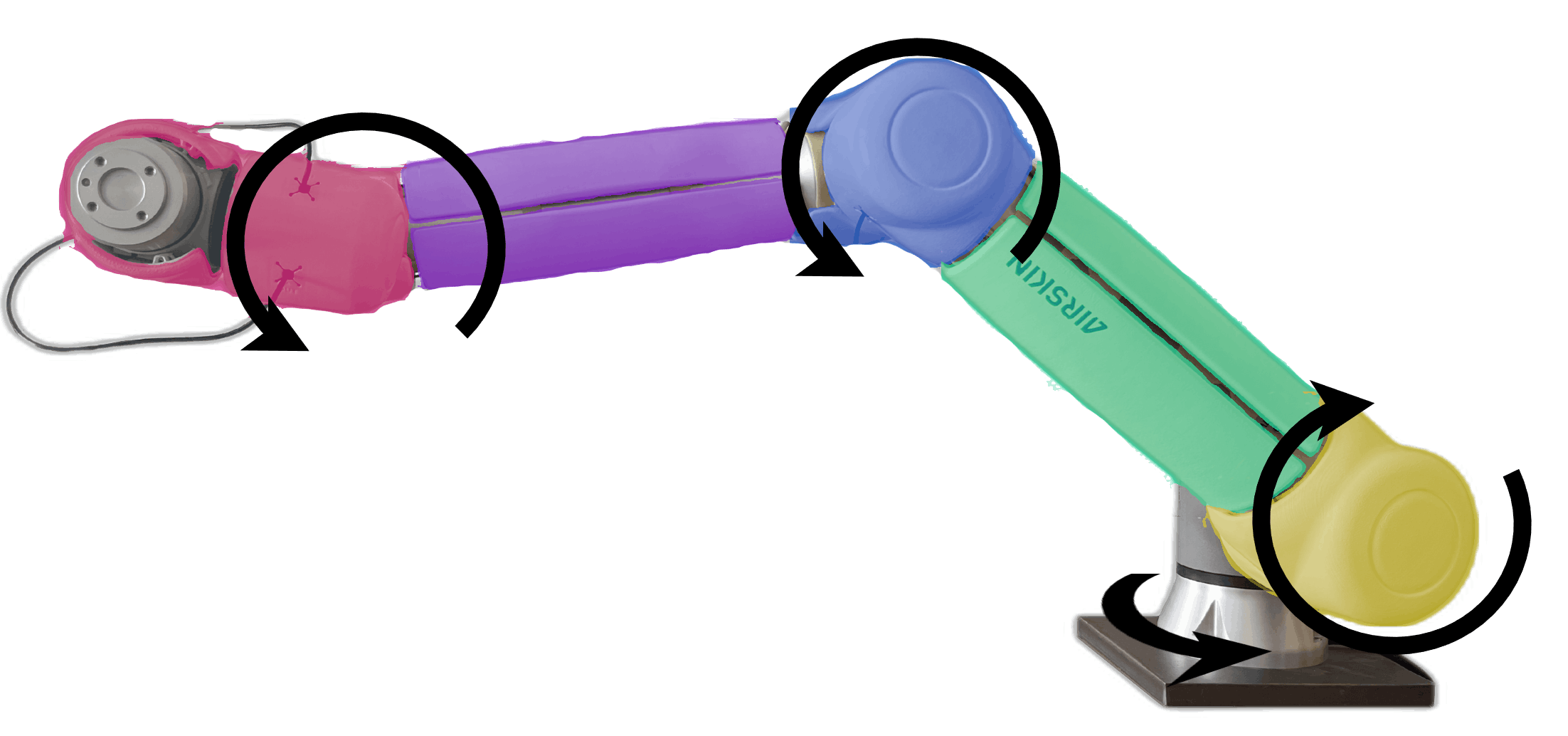}
    \includegraphics[width=0.375\columnwidth]{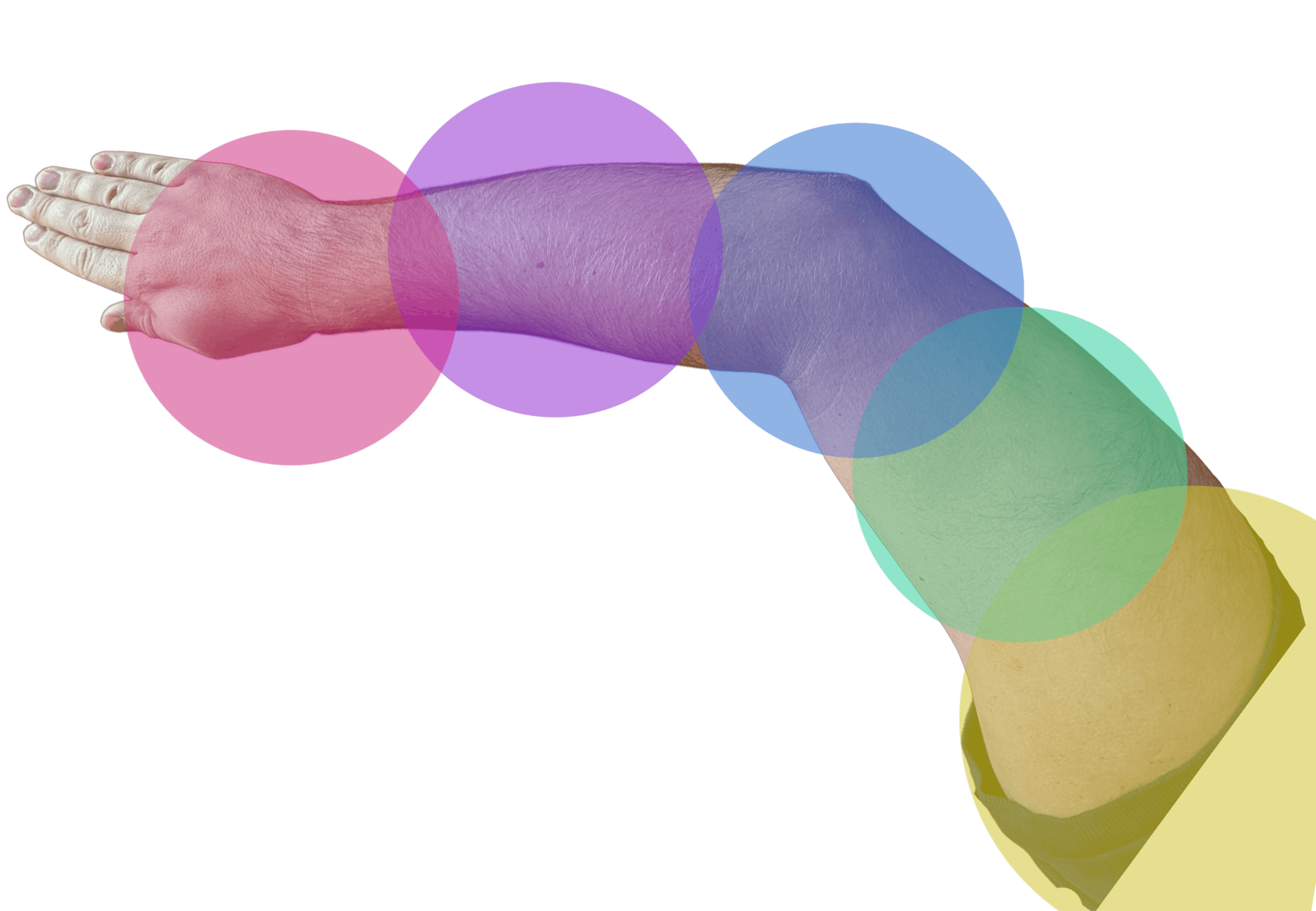}
    \caption{UR10e manipulator covered with ten AIRSKIN pressure pads compared to a human arm. Pads are grouped into shoulder (yellow), upper-arm (green), elbow (blue), forearm (purple), and wrist (pink) regions. The arrows correspond to the direction of movements in \acf{ur} and \acf{lr}.}
    \label{fig:platform}
\end{figure}

We utilized the Universal Robots UR10e 6-axis manipulator. Our robot is also equipped with the AIRSKIN whole-body sensitive skin. There are 10 pads that cover the whole surface of the robot. Robot links are covered by two semicylindrical pads; joints are covered with additional pads (Fig.~\ref{fig:platform}, left).  Each pad is a closed shell with a compressor and a pressure sensor inside, sending one pressure value. Therefore, it is possible to detect which pad is touched and what is the pressure change, but not exactly where on the given pad. We group pads into upper-limb analogues with one to three pads per body part. The pads are colored according to the body part in \figref{fig:platform}: shoulder (yellow), upper arm (green), elbow (blue), forearm (purple), and wrist (pink).

The whole system is implemented using \ac{ros} Noetic. Using a special AIRSKIN maintenance controller, the pressure values from individual pads can be retrieved at approximately \qty{25}{\hertz}. Unlike in the standard AIRSKIN deployment in the industry where any detected collision stops the robot, here we retrieve the pressure values and process them for our own collision detection, isolation, and identification. The values are further processed through a filtering node that provides changes in pressure with respect to a no-contact state. 
The robot is controlled in joint position or joint velocity mode, depending on the selected reflex. We provide the code at \censorurl{https://humanoids-ctu.github.io/Nociceptive-Reflexes}. We performed all of our experiments in the real world, but as the AIRSKIN technology is not open-source, the code contains a simulator of the UR10e with AIRSKIN to test the reflexes.

\subsection{Pressure-to-pain Gain}
Let $p$ denote the change (from a non-touch baseline) in pressure measured by the activated skin pad. The controller outputs a scalar gain $G(p)\in[0,1]$, where $G(p)=0$ means no withdrawal, and $G(p)=1$ means full withdrawal. 

\textbf{Discrete pain model.}
This model, $G_d(p)$, is based on the findings of \etal{Serrao}~\cite{serrao2006kinematic}. At \acp{pl} 1 and 2, no measurable joint displacement is produced and, therefore, the assignment is $G_d(p) = 0$. \ac{pl} 4 represents the maximum reflex response and is assigned to $G_d(p) = 1$. The gain in \ac{pl} 3 was determined by computing the ratio of joint displacements observed in \acp{pl} 3 and 4, producing a value of $G_d(p) = 0.35$. The final discrete pain model is defined as
\begin{equation}
    G_d(p) =
\begin{cases}
0, & p < 2\ \mathrm{hPa} \quad (\text{PL 1})\\
0, & 2 \leq p < 6\ \mathrm{hPa} \quad (\text{PL 2})\\
0.35, & 6 \leq p < 12\ \mathrm{hPa} \quad (\text{PL 3})\\
1, & p \geq 12\ \mathrm{hPa} \quad (\text{PL 4})
\end{cases}
    \label{discretePainGain}
\end{equation}
In terms of pressure change $p$ in the pad, the \acp{pl} are divided by the upper thresholds of $2, 6$, and \qty{12}{\hecto \pascal}. The values were selected empirically during preliminary experiments and are tightly bounded to our specific robot and skin.

\textbf{Continuous pain model.}
As biological pain intensity is graded, the implemented study used a continuous mapping defined as a sigmoid, $\sigma$ centered about the value of 0.5 as
\begin{equation}
\sigma(x;k)=\frac{1}{1+e^{-k(x-0.5)}},
\end{equation}
where $x\in[0,1]$ is the normalized pressure value and $k$ controls the steepness. We used $k=3$, as at this steepness, the sigmoid produces a gradual, monotonically increasing response consistent with the neural activation reported by \etal{Coghill}~\cite{pain_intensity_processing}, while its nonlinearity captures the graded perceptual pain response described by \etal{Price}~\cite{visual_analogue_scales}. The sigmoid is further scaled to provide the normalized $G(p)$ between 0 and 1 as
\begin{equation}
\label{eq:scaled_gain}
\sigma_s(x;k)=\frac{\sigma(x;k)-\sigma(0;k)}{\sigma(1;k)-\sigma(0;k)}
\end{equation}

With the continuous model (see Fig.~\ref{fig:schema}, top right), we utilize the pressure thresholds and \acp{pl} from the discrete model---we added threshold of \qty{20}{\hecto\pascal} for \ac{pl} 4 to allow for sigmoid computation. The gain is computed separately for each \ac{pl} range. Below the upper bound of \ac{pl} 2, no reflex is triggered and $G(p) = 0$. At \ac{pl} 3, the scaled sigmoid interpolates between $G(p) = 0$ and $G(p) = 0.35$. At \ac{pl} 4, it interpolates between $G(p) = 0.35$ and $G(p) = 1$. Everything above the last threshold is counted as $G(p) = 1$. This piecewise structure ensures continuity at the boundaries of individual \acp{pl} while retaining the biological foundation established in the discrete model. The continuous pain model with the given thresholds can be seen in the top right of \figref{fig:schema} and is mathematically defined as

\begin{equation}
G(p) =
\begin{cases}
0, & p < 6\ \mathrm{hPa}\\
0.35 \cdot \sigma_s\left(\frac{p - 6}{6}; k\right), & 6 \leq p < 12\ \mathrm{hPa}\\
0.35 + 0.65 \cdot \sigma_s\left(\frac{p - 12}{8}; k\right), & 12 \leq p < 20\ \mathrm{hPa}\\
1, & p \geq 20\ \mathrm{hPa}
\end{cases}
\label{eq:pain_lev_cont}
\end{equation}

\subsection{Reflexes}
We map human upper-limb joints to the manipulator as follows: shoulder internal rotation and adduction correspond to the base and shoulder joints ($J_{b}$, $J_{s}$), respectively. Elbow flexion to the elbow joint ($J_{e}$) and wrist movements to the wrist joints ($J_{w_i}; i \in [1,2,3]$). We only used the first joint of the wrist, which corresponds to wrist flexion.

We extracted the maximum displacement vector (range of motion) $\mathbf{q}_{\max}$ from the literature studying human range of motion~\cite{shoulder_flexion_rom, elbow_flexion_rom, wrist_rom}. 
The maximal joint ranges for the robot become: 
\begin{equation}
\mathbf{q}_{\max}=[40^\circ,\;70^\circ,\;146^\circ,\;73^\circ,\;0^\circ,\;0^\circ]^T.
\label{eq:qmax}
\end{equation}

The behavior of two of the reflexes is then derived from these ranges. The direction of these movements corresponds to shoulder internal rotation (we assume the manipulator is a right arm), shoulder adduction, elbow flexion, and wrist flexion.

\begin{figure}[h]
    \centering
    \includegraphics[width=1\columnwidth]{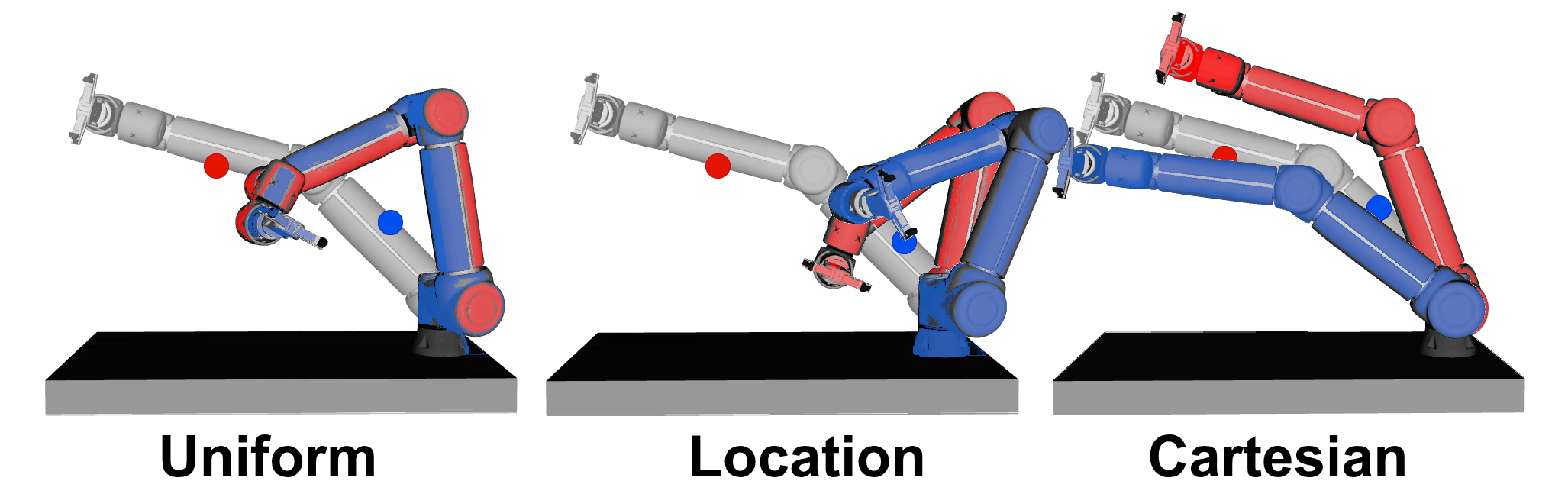}
    \caption{Illustration of the three different reflex types. Pain gain $G(p) = 1$. The initial pose is shown in grey, while the colored circles represent the points of touch and colored robots the corresponding reflexes. See the accompanying video for full movements.}
    \label{fig:reflexes_comp}
\end{figure}

Reflexes are triggered only for pad pressure level changes above 6 hPa (\equationref{eq:pain_lev_cont}). Schematic illustration of their behavior is in \figref{fig:reflexes_comp}.

\textbf{\ac{ur}.} When triggered by above-threshold contact anywhere on the arm, this reflex applies a fixed increment to the current position of the first four joints of the manipulator. We selected the movement range to correspond to $G(p) = 0.5$, i.e., to be a moderate response to a contact. Thus, the displacement vector is calculated as
\begin{equation}
\mathbf{q}_{\mathrm{ref}}=\frac{1}{2}\mathbf{q}_{\max} =
\begin{bmatrix}
     20^\circ,\;35^\circ,\;73^\circ,\;36.5^\circ,\;0^\circ,\;0^\circ
\end{bmatrix}^T.
\label{eq:uniform}
\end{equation}
This displacement vector is added to the current pose of the robot, checked against individual joint limits and clipped if necessary. Joint position control was used.

\textbf{\acf{lr}.} The biologically inspired joint-space behavior scales the maximum displacement by the pain gain $G(p)$ and by a diagonal location-dependent weighting matrix $\mathbf{W}(p,\ell)$ as
\begin{equation}
\mathbf{q}_{\mathrm{ref}}=G(p)\,\mathbf{W}(p,\ell)\,\mathbf{q}_{\max},
\label{eq:location}
\end{equation}
where $\ell$ is the activated body region and $\mathbf{W}(p,\ell) \in \mathbb{R}^{6 \times 6}$ contains weights for individual joints $j \in \{b,s,e,w\}$ of the base, shoulder, elbow, and wrist (the same weight is used for all three joints of the wrist). The weights were selected from the upper-limb \ac{nwr} literature and refined by manual testing for visually plausible UR10e motions. Table~\ref{tab:weightingMatrix} provides an overview of the weights used in our implementation, together with the biological explanation of the movement. The reflex is dominated by shoulder adduction and elbow flexion, with involvement of the base of wrist flexion depending on the contact location.

The displacement vector is then added to the current pose as in the case of \ac{ur}, with the same control scheme.

\begin{table*}[t]
\centering
\footnotesize
\caption{Spatial organization of the human \ac{nwr} and corresponding joint-weight selection for \acf{lr}. Weights $w_{J_j}$ correspond to weights of individual joints $j \in \{b,s,e,w\}$ of base, shoulder, elbow and wrist (only one weight is used for all three wrist joints).}
\label{tab:weightingMatrix}
\begin{tblr}{
width=\textwidth,
colspec={
Q[c,0.75cm]
Q[l,2.5cm]
X[4.2,l]
Q[c,2.8cm]
},
row{1} = {font=\bfseries},
rows = {valign=m},
rowsep = 2pt,
colsep = 3pt,
hline{1,Z} = {1.1pt, black},
hline{2} = {0.8pt, black},
hline{5,6,9,10,11} = {0.3pt, black},
}

Pain &
Stimulus location &
Description / Interpretation &
\textbf{Weights} \newline $[w_{J_b}, w_{J_s}, w_{J_e}, w_{J_w}]$ \\

1--2 &
Any &
Sensory nociceptive processing without mechanical withdrawal. No joint motion occurs, but awareness and compliance are maintained. &
$[0, 0, 0, 0]$ \\

\hline[1.2pt]
\SetCell{bg=gray!10} 3 &
\SetCell[c=3]{l,bg=gray!10,font=\itshape} Local protective adjustment without full withdrawal &
&
\\
\hline[0.8pt]

&
Shoulder / upper arm &
Primarily excites shoulder anteflexion, supported by elbow flexion and occasional shoulder adduction. &
$[0.2, 0.8, 0.6, 0.0]$ \\

&
Elbow / forearm &
Elbow-dominated withdrawal with shoulder assistance and minor wrist response. &
$[0.0, 0.6, 0.8, 0.2]$ \\

&
Wrist &
Local wrist flexors are recruited, with dominant elbow flexion and supportive shoulder anteflexion to increase limb--stimulus distance. &
$[0.0, 0.3, 0.6, 0.8]$ \\

\hline[1.2pt]
\SetCell{bg=gray!10} 4 &
\SetCell[c=3]{l,bg=gray!10,font=\itshape} Whole-arm withdrawal with emphasis on flexion of the stimulated body part &
&
\\
\hline[0.8pt]

&
Shoulder / upper arm &
Proximal withdrawal reflex characterized by strong shoulder anteflexion and supporting elbow flexion. Depending on the precise stimulus location, shoulder adduction may also occur. Wrist involvement is minimal or absent. &
$[0.5, 1.0, 0.7, 0.1]$ \\

&
Elbow / forearm &
Elbow-dominant withdrawal reflex with significant shoulder participation and moderate wrist motion. &
$[0.0, 0.8, 1.0, 0.4]$ \\

&
Wrist &
Strong wrist flexion and adduction combined with pronounced elbow flexion and shoulder anteflexion to rapidly withdraw the hand. &
$[0.0, 0.6, 0.8, 1.0]$ \\
\end{tblr}
\end{table*}

\textbf{\ac{cr}.} The Cartesian reflex used the manipulator Jacobian to withdraw the contacted skin pad in the operational space along the surface normal away from the contact. Let $\mathbf{n}$ be the activated pad normal transformed into the robot base frame. The desired translational velocity is
\begin{equation}
\mathbf{v}_c=-v_{\max}\,G(p)\,\mathbf{n},
\label{eq:cart_vel}
\end{equation}
where $v_{\max}$ is a velocity scale. The negative sign moves the pad away from the contact---the normals of the pads are always pointing away from the surface and are pre-defined in local frames. 

Joint velocities are computed using resolved-rate control~\cite{whitney1969resolved} as
\begin{equation}
\dot{\mathbf{q}}_{ref}=\mathbf{J}^{\dagger}(\mathbf{q})
\begin{bmatrix}
\mathbf{v}_c \\ \mathbf{0}
\end{bmatrix},
\label{eq:rmrc}
\end{equation}
where $\mathbf{J}^{\dagger}$ is the pseudoinverse of the Jacobian from the robot base to the contacted pad. The resulting joint velocities are streamed for a fixed reflex duration (0.5 s).

\textbf{Comparison.} The uniform reflex is stereotypical---the same joint increment independent of which robot part was touched---and hence predictable. The location-dependent reflex is strongly biologically inspired and the behavior is modulated by the pain gain and the contact location. The Cartesian reflex constitutes an engineering solution: the contacted skin pad will be moved away in Cartesian space, employing all robot joints according to the mapping through the Jacobian. Visual comparison of reflexes can be seen in \figref{fig:reflexes_comp} and in the accompanying video.

\subsection{State Machine}
During the experiments, the robot was controlled using a state machine with three states: 
\begin{enumerate*}[label=(\textbf{\Roman*})]
    \item TASK;
    \item REFLEX;
    \item and RETURNING.
\end{enumerate*}
During TASK, the robot performs the pre-defined trajectory. At the same time, it senses and processes skin data. If nonzero pain gain is detected, the robot saves the current pose, pauses trajectory execution, and switches to REFLEX---if there is a new or stronger contact during REFLEX, the trajectory and duration can be updated. After execution, it switches to RETURNING and returns to the saved configuration and continues TASK. See \figref{fig:schema} for a visual representation.

%%%%%%%%%%%%%%%%%%%%%%%%%%%%%%%%%%%%%%%%%%%%%%%%%%%%%%%%%%%%%%%%%%%%%%%%%%%%%%%%
\section{Results}

We prepared the interaction with the robot as a small game for the participants. In each condition, the robot performed the same trajectory representing the boundaries of an $\infty$ sign---see the accompanying video. The task of the participants was to touch the skin pad that changed color from blue to red.
Participants were asked to touch the center of the active pads with one finger using varying pressure levels and observe the robot reaction. We conducted a within-subject study with three conditions: \acf{ur}, \acf{lr}, \acf{cr}. The order of the conditions was counterbalanced in three orders, with five participants per order. Each condition lasted approximately five minutes and involved at least ten touches. After each condition, participants completed per-condition questionnaires. After all conditions, they completed forced-choice comparative questions and open-ended questions. The entire session lasted approximately 25 minutes. The questionnaire was given in English for all participants. Participants provided written informed consent. The study was approved by the \xblackout{Committee for Research Ethics at the Czech Technical University in Prague (reference 0000-20/25/51902/EKCVUT)}.

\textbf{Evaluation and Metrics.} Per-condition ratings included four Godspeed subscales~\cite{bartneck2009}: Anthropomorphism, Animacy, Likeability, and Perceived Safety. 
And eight custom 5-point Likert scale~\cite{sullivan2013analyzing} items grouped into 3 categories asking users to score the following statements:
\begin{itemize}
    \item Naturalness and Realism:
    \begin{itemize}
        \item The robot’s reaction to my touch felt natural;
        \item The robot’s withdrawal movement made sense to me.
    \end{itemize}
    \item Safety \& Comfort:
    \begin{itemize}
        \item The robot’s response speed felt appropriate;
        \item I felt in control of the interaction at all times;
        \item I would feel comfortable working near this robot.
    \end{itemize}
    \item Reflex Perception
    \begin{itemize}
        \item The robot’s movement felt proportional to how hard I pressed;
        \item The robot’s behavior reminded me of a human;
        \item The robot reacted as if it was avoiding discomfort or damage.
    \end{itemize}
\end{itemize}
The final comparative questions asked which condition felt the most natural, most human-like, safest, and most appropriate as a touch reaction. The users also had the opportunity to fill the following open questions:
\begin{itemize}
    \item What did you like most about the robot’s behavior?
    \item What felt unnatural or uncomfortable during the interaction?
    \item Please explain your preference (or lack of preference) between the conditions.
    \item Do you have suggestions for improving the robot’s reactions?
\end{itemize}
The complete questionnaire and answers are available at \censorurl{https://humanoids-ctu.github.io/Nociceptive-Reflexes}.

We used descriptive statistics, Friedman tests to analyze repeated-measures differences, and pairwise Wilcoxon signed-rank tests as post-hoc analyses when the Friedman test yielded significant results.

\textbf{Participant characteristics.} We evaluated the study with 15 participants (8 male, 7 female; ages 20--51, median 26, mean 27.9). 
The initial screening showed that 7\% never interact with robotic systems, 53\% rarely, 33\% monthly, and 7\% weekly or more. In the specific case of the robotic arm, 60\% never had an experience with it, 20\% had brief experience and 20\% had extensive experience. On a five point scale from very negative to very positive, the average attitude towards robot was 4.4, with the lowest grade of 3. 

\subsection{Godspeed Subscales}
\begin{figure}[tb]
    \centering
        \begin{minipage}{1\columnwidth}
            \begin{center}
                \begin{minipage}{0.49\textwidth}
                    \centering
                    \includegraphics[width=1\textwidth]{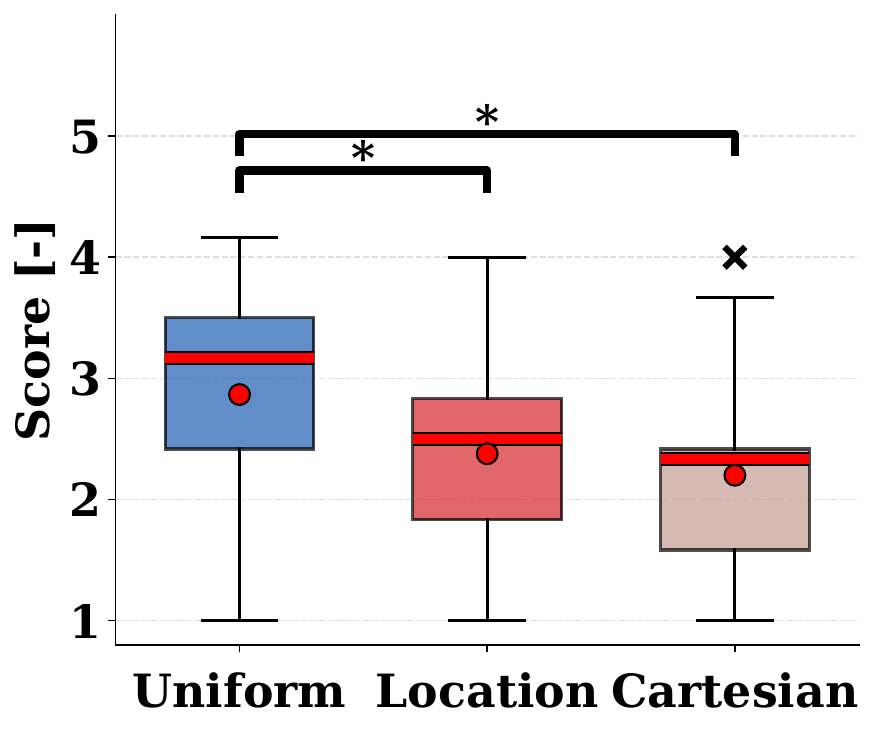}\\
                    {\footnotesize (a) \textbf{Anthropomorphism}}
                \end{minipage}
                \begin{minipage}{0.49\textwidth}
                    \centering
                    \includegraphics[width=1\textwidth]{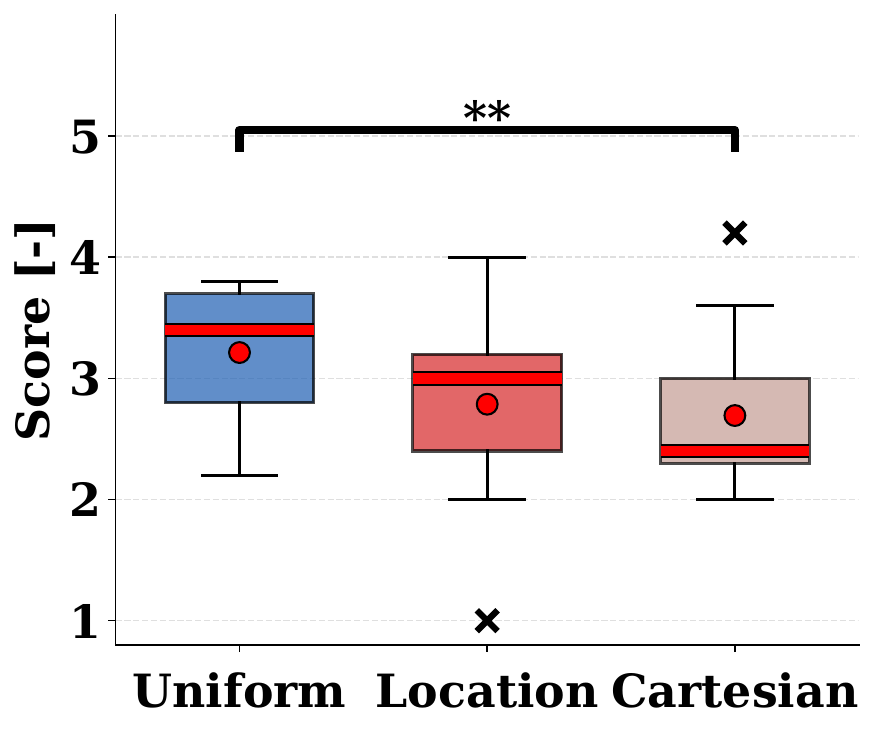}\\
                     {\footnotesize (b) \textbf{Animacy} }
                 \end{minipage}
            \end{center}
        \end{minipage}

        \begin{minipage}{1\columnwidth}
            \begin{center}
                \begin{minipage}{0.49\textwidth}
                    \centering
                    \includegraphics[width=1\columnwidth]{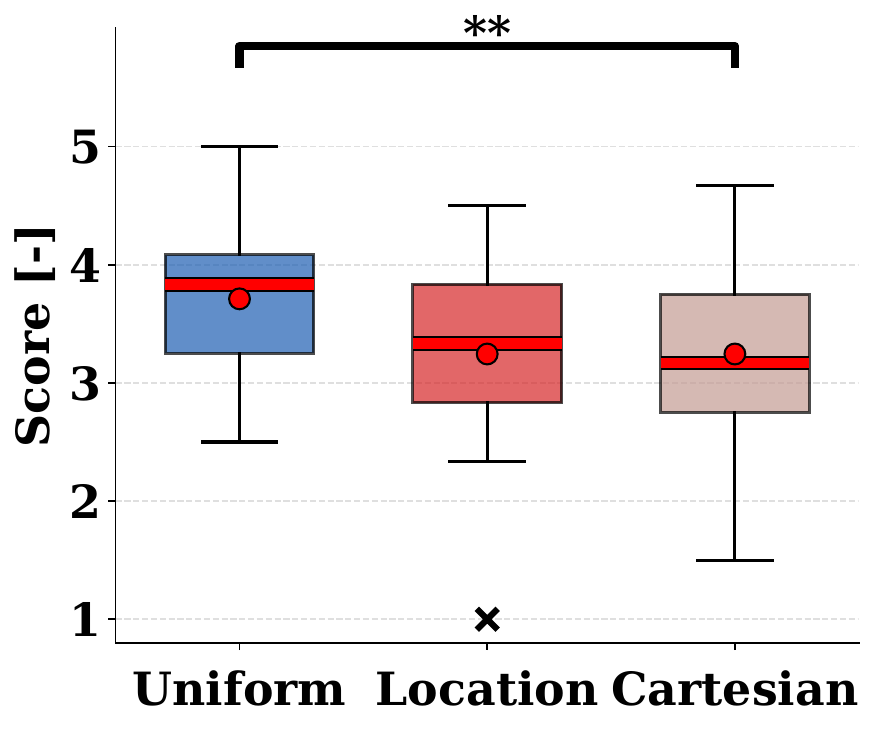}\\
                    {\footnotesize (c) \textbf{Likeability} }
                \end{minipage}
                 \begin{minipage}{0.49\textwidth}
                    \centering
                    \includegraphics[width=1\columnwidth]{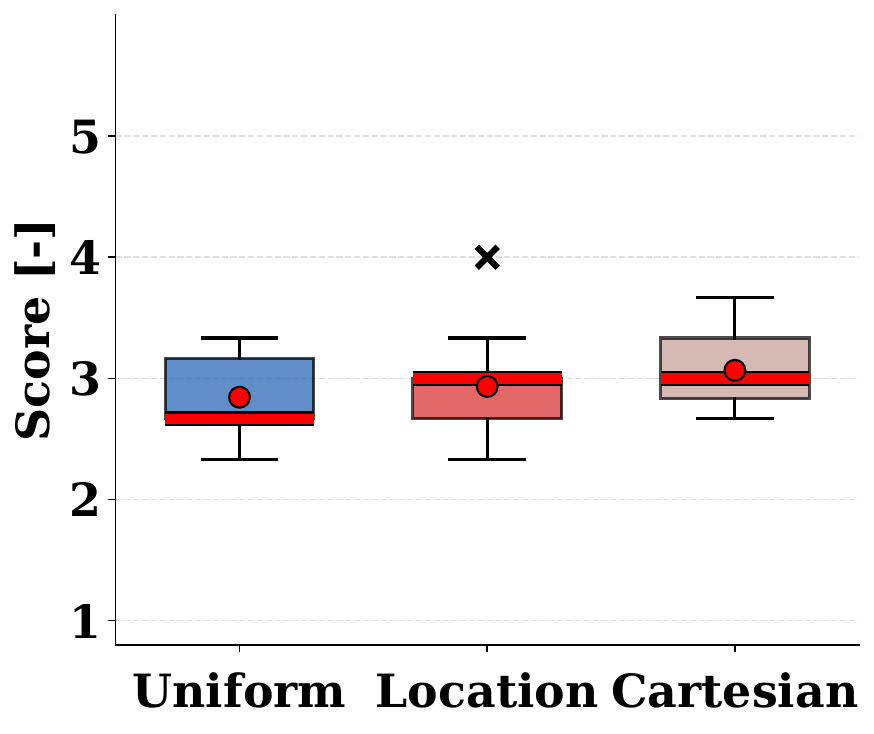}\\
                    {\footnotesize (d) \textbf{Perceived Safety}}
                \end{minipage}
           \end{center}
        \end{minipage}

    \caption{Godspeed scores per condition for each of the four subscales. The red points represent mean and red lines median. Boxes are from 25th to 75th percentile with whiskers showing extreme non-outlier point and crosses showing outliers. Statistical significance was evaluated using the Wilcoxon signed-rank test, where *, **, *** denote p-values below 0.05, 0.01, and 0.001, respectively.}
    \label{fig:godspeed_scores}
\end{figure}

The first evaluation was done using the Godspeed questions. The boxplots results are shown in \figref{fig:godspeed_scores}. The \acf{ur} scored the best in Anthropomorphism, Animacy, and Likeability. The difference is significant w.r.t the \acf{cr} in all three cases; in Anthropomorphism, the difference is also significant w.r.t \acf{lr}. 
In the last subscale---Perceived Safety---there is almost no difference in the appreciation of the three different reflexes. 

\subsection{Custom Subscales}
\begin{figure}[htb]
    \centering
        \begin{minipage}{1\columnwidth}
            \begin{center}
                \begin{minipage}{0.49\textwidth}
                    \centering
                    \includegraphics[width=1\textwidth]{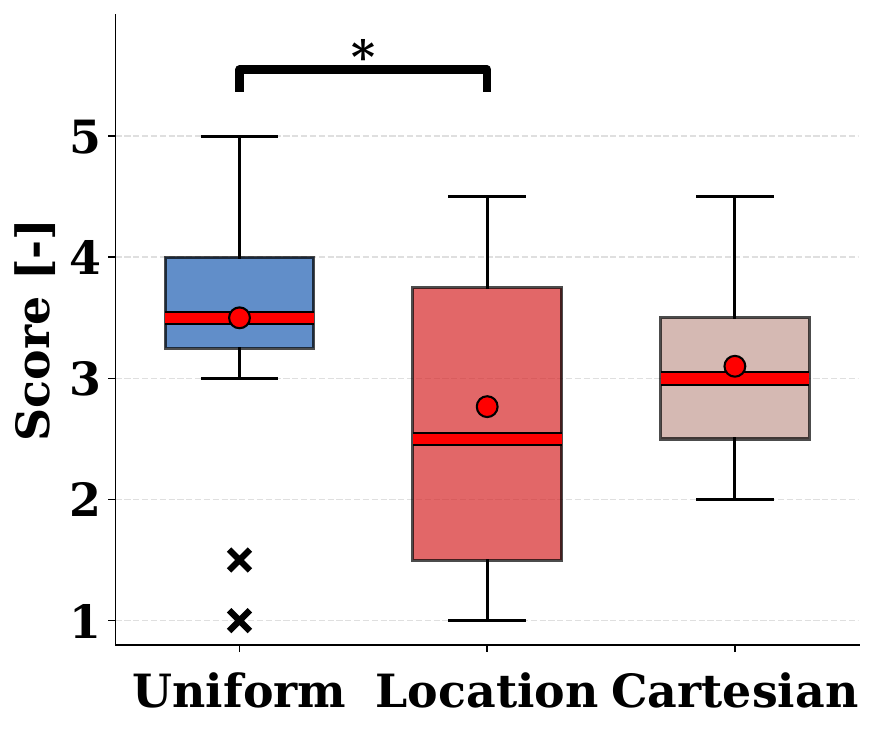}\\
                    {\footnotesize (a) \textbf{Naturalness \& Realism}}
                \end{minipage}
                \begin{minipage}{0.49\textwidth}
                    \centering
                    \includegraphics[width=1\textwidth]{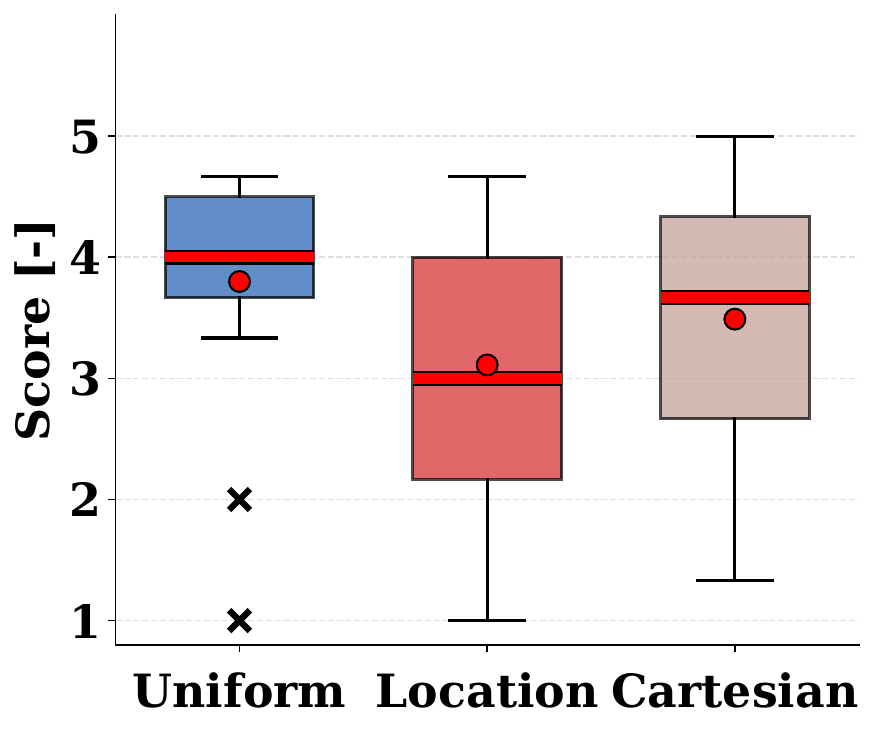}\\
                     {\footnotesize (b) \textbf{Safety \& Comfort} }
                 \end{minipage}
            \end{center}
        \end{minipage}

        \begin{minipage}{1\columnwidth}
            \begin{center}
                \begin{minipage}{0.49\textwidth}
                    \centering
                    \includegraphics[width=1\columnwidth]{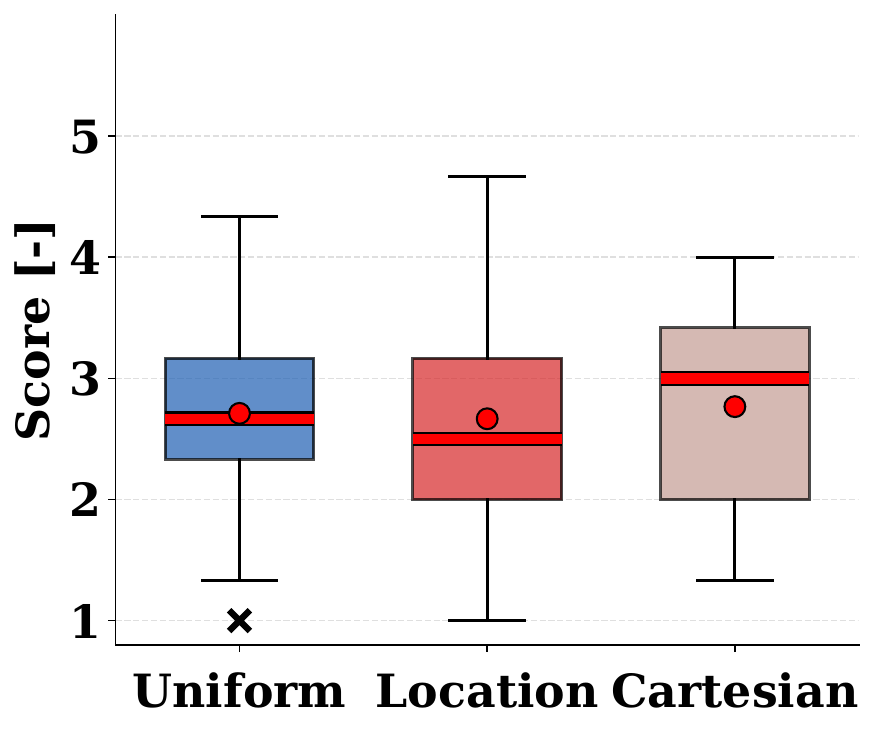}\\
                    {\footnotesize (c) \textbf{Perception} }
                \end{minipage}
           \end{center}
        \end{minipage}

    \caption{Custom scales Likert scores per condition. The red points represent mean and red lines median. Boxes are from 25th to 75th percentile with whiskers showing extreme non-outlier point and crosses showing outliers. Statistical significance was evaluated using the Wilcoxon signed-rank test, where *, **, *** denote p-values below 0.05, 0.01, and 0.001, respectively.}
    \label{fig:custom_scores}
\end{figure}

We further designed custom subscales to assess whether the movement felt natural (Naturalness \& Realism; \figref{fig:custom_scores} (a)), whether the speed was appropriate and whether they felt in control (Safety \& Comfort \figref{fig:custom_scores} (b)), and whether the movement was proportional to their touch and reminiscent of human behavior (Perception; (\figref{fig:custom_scores} (c)).

The results suggest several conclusions. Firstly, \ac{ur} is perceived as more natural (with a statistically significant difference from \ac{lr}) and safe, but its speed/trajectory is not perceived as proportional to the touch. Secondly, \ac{lr} has the highest variance in all three subscales. 

\subsection{Comparative Questions}
\begin{figure}[h]
    \centering
    \includegraphics[width=0.85\columnwidth]{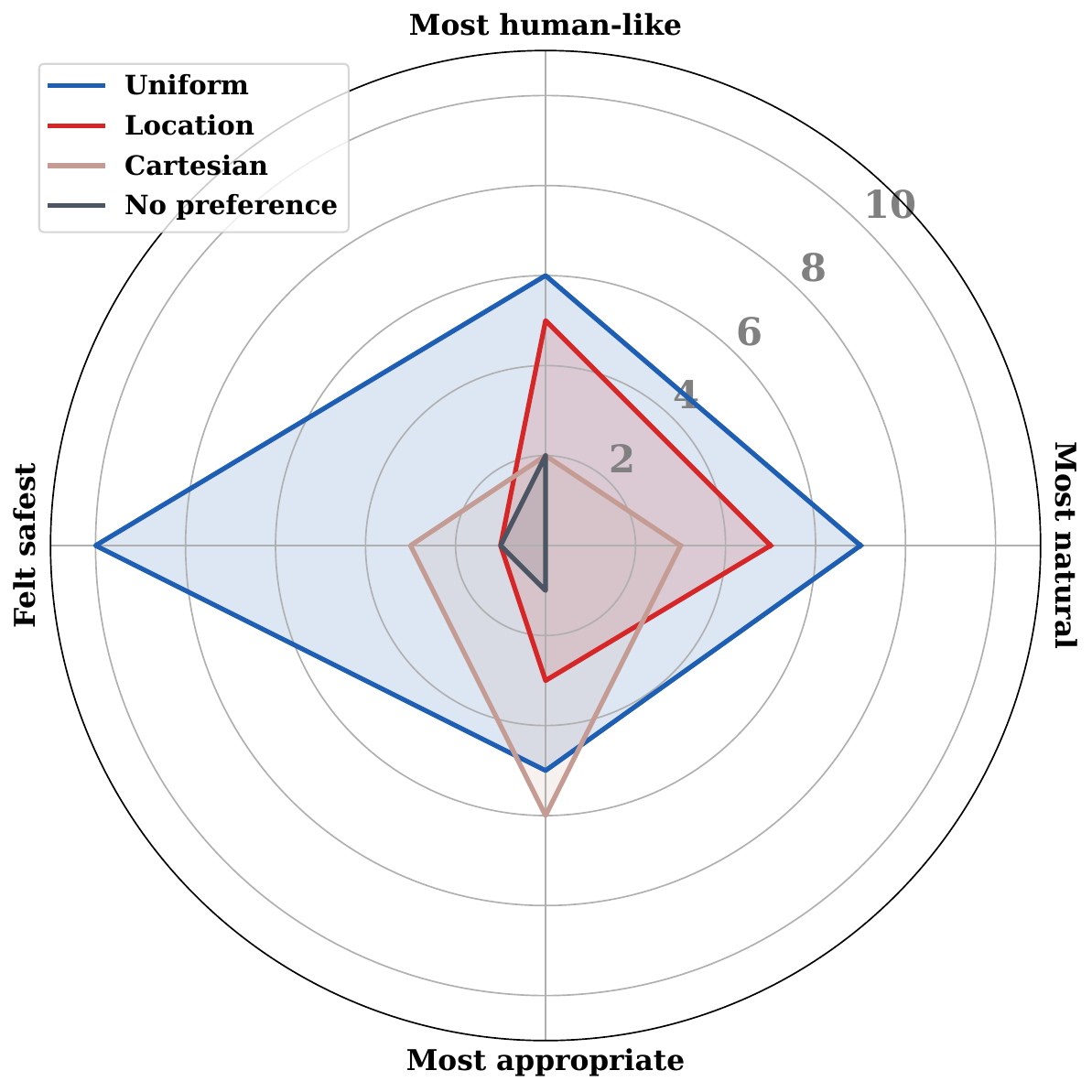}
    \caption{Radar plot of participant preferences across conditions for naturalness, human-likeness, safety, and reaction appropriateness. The numbers represent number of votes (out of 15 participants) for the given option.}
    \label{fig:comparative}
\end{figure}

After completing all three conditions, we also asked the participants comparative questions. They could select one of the conditions or select the fourth option of no preference. The results are visualized in \figref{fig:comparative}. Interestingly, \ac{ur} that was meant as a baseline performed the best in human-like and natural movements with the difference of 1 and 2 votes from the second best \ac{lr}. Note that the difference is small and with our sample size it should be interpreted with caution. On the other hand, \ac{ur} felt the safest with 7 more votes on the second best condition. This result suggests that this simple reflex may be the best for \ac{hri}. We believe that this is caused by the fact that this condition is the most predictable. 
As the most appropriate reflex, the participants selected \ac{cr}---with a small margin of 1 vote.

\subsection{Qualitative feedback}

The answers to open-ended questions were consistent with the previous sections of the results. Comments about the uniform reflex were dominated by predictability and controllability. Participants reported that the robot's action was easy to learn and that this made the interaction feel safer, even though some described the motion as repetitive or less expressive. 

The feedback on the bio-inspired location-dependent behavior was polarized. Some participants interpreted the variability as more lifelike because the robot did not always repeat the same motion. Others interpreted the same variability as uncertainty. The most important criticism was directional: in certain postures, the selected joint combination could cause part of the robot to move toward the participant while another part withdrew. 

The Cartesian behavior was easier to explain after observation. The participants recognized that the touched pad moved away from the contact, which likely explains why it received the highest number of votes for the most appropriate reaction to touch. Its main negative feedback was about abruptness. The avoidance movement was applied directly, without smoothing the trajectory in task space or joint space. 

%%%%%%%%%%%%%%%%%%%%%%%%%%%%%%%%%%%%%%%%%%%%%%%%%%%%%%%%%%%%%%%%%%%%%%%%%%%%%%%%
\section{Discussion, Conclusion and Future Work}
In this work, we presented an artificial nociceptive reflex pipeline for a manipulator equipped with whole-body tactile sensing. By mapping extrinsic pressure changes to a scalar pain gain using a biologically-informed continuous model, we implemented and compared three distinct withdrawal strategies: a baseline \acf{ur}, a biologically-inspired \acf{lr}, and an engineered \acf{cr}. The study was conducted as a game that required users to touch a part of the robot colored red. A within-subject human evaluation with 15 participants was conducted to assess the perceived naturalness, safety, and appropriateness of these reflexes during direct physical human-robot interaction.  Data from experiments (the study questionnaire and responses), code, and video recordings are available at \censorurl{https://humanoids-ctu.github.io/Nociceptive-Reflexes}.

Arguably, the most important finding is the trade-off between biological fidelity, geometric legibility, and predictability. The \ac{lr} strategy best reflects the upper-limb \ac{nwr} literature---stimulus location and pain level determine which joints participate. However, this did not translate into the highest perceived naturalness or safety. 

The embodiment gap between the robot manipulator and a human arm is probably too large for participants to subconsciously relate the robot movements to humans. 
The behavior of \ac{ur} provided a consistent response that participants could predict after a few touches. This is consistent with the literature on perceived-safety in physical \ac{hri}, where predictability, controllability, and transparency are central factors in how safe a robot feels~\cite{rubagotti2022perceived,tusseyeva2022perceived}. In both \ac{lr} and \ac{ur} joint movements were ``toward the body'', which is in line with the biological reflexes, but in some configurations could results in movement of the the touched pad toward and not away from the contact. 
Instead, the \ac{cr} behavior always moves the contacted surface away. This should be considered in particular in physical HRI contexts, where safety is more important than perceived safety.
The \ac{cr} is also more transferable across morphologies than a manually chosen joint-weight table. It was ranked as the most appropriate reaction to touch. 
The feedback questions later revealed that the participants were mostly concerned about the motion and velocity profiles of this behavior. The behavior was perhaps too abrupt and brief in duration. A smoother and longer movement could be better perceived. 
In the future, we would like to repeat the experiments with different robot morphologies (e.g. with a humanoid robot) or introduce different motions using, for example, recorded human motions or directly applying motion retargeting frameworks~\cite{klein2022riemannian}.

We evaluated a complete artificial nociceptive framework on a real robot with whole-body skin and performed a controlled within-subject study. However, the users pointed out some possible improvements. Firstly, even though the robot's responsiveness was marked positively, the difference between light and firm touch was not always perceivable. This comes from the characteristics of the skin used. It is sending data with the frequency of about \qty{25}{\hertz}, which is enough for responsive behavior, but pressure-to-force mapping is difficult---it depends not only on the touch force, but also on contact area, robot velocity, or touch location. Therefore, the same touch could result in different pain gains. This is paired with another suggestion of making the intent of the robot better visible to the participants, e.g., by showing the touch intensity by colors of LED lights on the skin. This is unfortunately not possible with AIRSKIN as changing color actually increases latency. However, these problems could be solved with the use of different artificial skins or other sensing devices.

%%%%%%%%%%%%%%%%%%%%%%%%%%%%%%%%%%%%%%%%%%%%%%%%%%%%%%%%%%%%%%%%%%%%%%%%%%%%%%%% 
\bibliographystyle{IEEEtran}
\bibliography{refs}

\end{document}